\documentclass[lettersize,journal]{IEEEtran}
\usepackage{amsmath,amsfonts}
\usepackage{algorithmic}
\usepackage{algorithm}
\usepackage{array}
\usepackage[caption=false,font=normalsize,labelfont=sf,textfont=sf]{subfig}
\usepackage{textcomp}
\usepackage{stfloats}
\usepackage{url}
\usepackage{verbatim}
\usepackage{graphicx}
\usepackage{cite}
\usepackage{orcidlink}
\usepackage{color}
\usepackage{booktabs}
\usepackage{multirow}
\usepackage[table]{xcolor}
\usepackage{pifont}

\definecolor{ourscolor}{gray}{0.9}
\newcommand{\ours}[1]{\cellcolor{ourscolor}{#1}}

\begin{document}

\title{Latent Bias Alignment for High-Fidelity Diffusion Inversion in Real-World Image Reconstruction and Manipulation}

\author{
    Weiming Chen$^{\orcidlink{0000-0002-0586-1278}}$,~\IEEEmembership{Student Member,~IEEE}, 
    Qifan Liu$^{\orcidlink{0000-0003-4949-1966}}$, 
    Siyi Liu$^{\orcidlink{0009-0001-4263-535X}}$, 
    Yushun Tang$^{\orcidlink{0000-0002-8350-7637}}$, 
    Yijia Wang$^{\orcidlink{0009-0004-9627-0677}}$,
    Zhihan Zhu$^{\orcidlink{0009-0001-1590-5318}}$,
    and Zhihai He$^{\orcidlink{0000-0002-2647-8286}}$,~\IEEEmembership{Fellow,~IEEE}
\thanks{Weiming Chen, Siyi Liu, Yijia Wang, Zhihan Zhu, and Zhihai He are with the Department of Electrical and Electronic Engineering, Southern University of Science and Technology, Shenzhen 518055, China. 

Qifan, Liu is with Undergraduate School of Artificial Intelligence, Shenzhen Polytechnic University, Shenzhen 518055, China.

Yushun Tang is with Huawei Technologies Co., Ltd., Shenzhen, China.

Zhihai He is also with the Pengcheng Lab, Shenzhen, China. }
\thanks{Corresponding author: Zhihai He (hezh@sustech.edu.cn).}}

\markboth{Manuscript under review}%
{Shell \MakeLowercase{\textit{et al.}}: A Sample Article Using IEEEtran.cls for IEEE Journals}


\maketitle

\begin{abstract}
Recent research has shown that text-to-image diffusion models are capable of generating high-quality images guided by text prompts. But can they be used to generate or approximate real-world images from the seed noise? This is known as the diffusion inversion problem, which serves as a fundamental building block for bridging diffusion models and real-world scenarios. However, existing diffusion inversion methods often suffer from low reconstruction quality or weak robustness. Two major challenges need to be carefully addressed: (1) the misalignment between the inversion and generation trajectories during the diffusion process, and (2) the mismatch between the diffusion inversion process and the VQ autoencoder (VQAE) reconstruction. To address these challenges, we introduce a latent bias vector at each inversion step, which is learned to reduce the misalignment between inversion and generation trajectories. We refer to this strategy as Latent Bias Optimization (LBO). Furthermore, we perform an approximate joint optimization of the diffusion inversion and VQAE reconstruction processes by learning to adjust the image latent representation, which serves as the connecting interface between them. We refer to this technique as Image Latent Boosting (ILB). Extensive experimental results demonstrate that the proposed method significantly improves the image reconstruction quality of the diffusion model, as well as the performance of downstream tasks, including image editing and rare concept generation.
\end{abstract}


\section{Introduction}

\IEEEPARstart{R}{ecent} advancements in text-to-image diffusion models have revolutionized the field of image synthesis~\cite{Ho2020DDPM,Ramesh2022DALL-E2,Saharia2022Imagen,Rombach2022LDM,Balaji2023eDiff-I,Betker2023DALL-E3,Podell2024SDXL,Sauer2024SDXLTurbo,Chen2024PixArt-alpha,Esser2024RectifiedFlowTransformers} and driven various downstream applications. Text-to-image diffusion models generate high-quality images by gradually transforming random noise samples into images guided by user-provided textual prompts. Latent Diffusion Models (LDMs)~\cite{Rombach2022LDM} train Diffusion Probabilistic Models (DPMs)~\cite{Ho2020DDPM} in a lower-dimensional representational space by utilizing a pre-trained VQAE and incorporating cross-attention layers into the model architecture, thereby turning diffusion models into powerful and flexible image generators. 
Recent studies have also demonstrated the strong potential of diffusion models in realistic image restoration~\cite{Zhang2025SSP-IR,Song2025Torch}, image editing~\cite{Hertz2023P2P,Cao2023MasaCtrl,Tumanyan2023PnP,Ju2024PIEBench}, subject-driven generation~\cite{Chen2024DisenDreamer,Zhao2025CatVersion,Wang2025MaskTI-TextReg}, and rare concept generation~\cite{Samuel2024SeedSelect,Samuel2023NAOSeedSelect}, further highlighting the importance of accurately bridging real images and generative diffusion models.
A critical challenge lies in mapping a real-world image to its corresponding seed noise, which can be used to reconstruct the image when combined with an appropriate textual prompt. Inversion is the technique that connects the generation model with real-world images, which has been widely studied in the GANs literature~\cite{Abdal2019Image2StyleGAN,Abdal2020Image2StyleGANplusplus,Gu2020mGANprior,Tov2021encoder4editing,Alaluf2021ReStyle}.

\begin{figure}[t]
\centering
\includegraphics[width=\columnwidth]{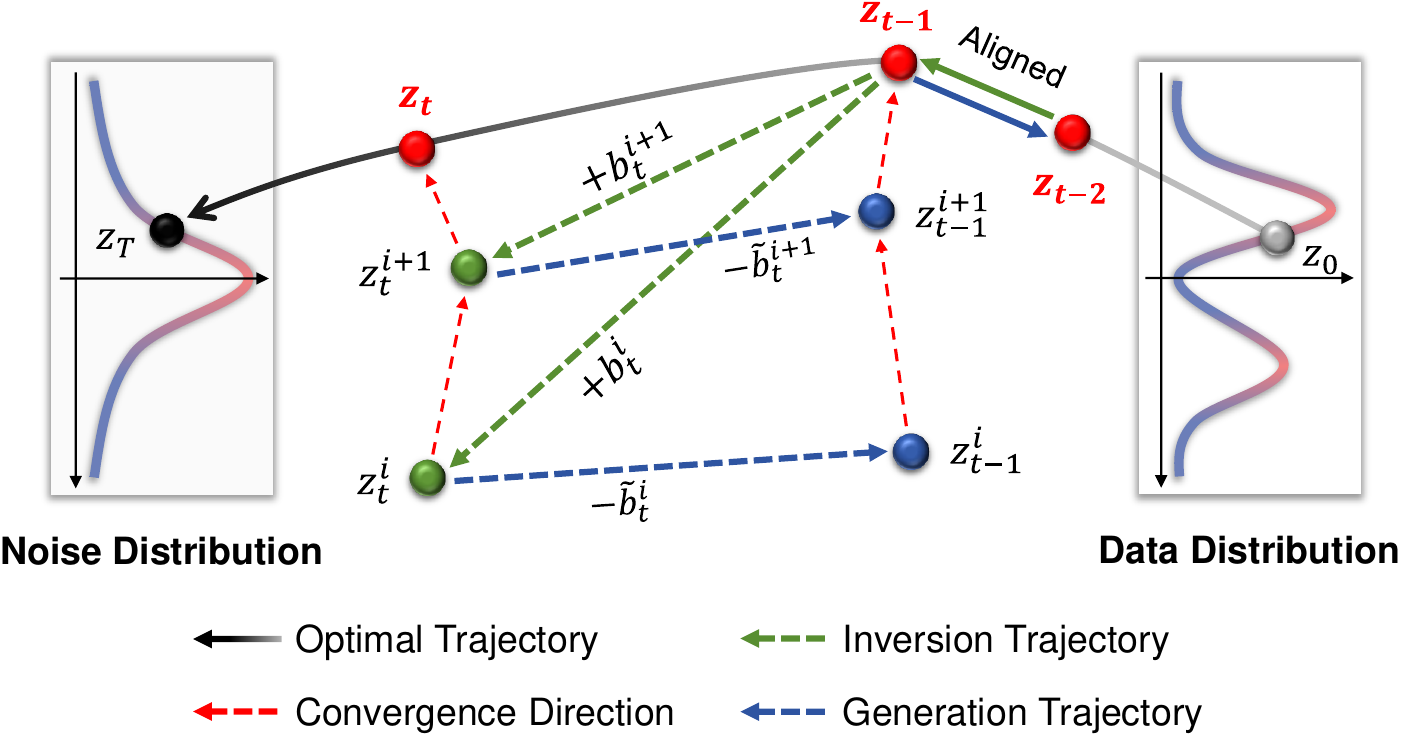}
\caption{The schematic illustration of the proposed Latent Bias Optimization (LBO) strategy, which learns the latent bias vector at each inversion step to align the inversion and generation trajectories, thereby seeking an optimal inversion trajectory for the given real-world image.}
\label{fig: LBO}
\end{figure}

Inversion in diffusion models aims to identify the appropriate combination of seed noise and conditional prompt embedding for the given image, making it a crucial step in enabling diffusion models to manipulate real-world images. Sampling from diffusion models can alternatively be viewed as solving the corresponding diffusion ODEs~\cite{Lu2022DPMSolver}. Inversion is, therefore, feasible under the assumption that the ODE process can be reversed in the limit of small steps. However, approximation errors can lead to misalignment between the inversion and generation trajectories. All existing diffusion inversion methods rely on this assumption and utilize various strategies to mitigate the degradation of reconstruction performance. Although they achieve some improvements, these methods either introduce burdensome additional parameters or suffer from poor stability due to the limitations of numerical iteration. We ask: \textit{Is it possible to build a reversible diffusion process in which the inversion and generation trajectories are aligned?} 

LDMs consist of two core components: a VQAE model for the image-latent domain transition and a diffusion UNet model for noise prediction. A mismatch between the diffusion inversion process (using the UNet model) and VQAE reconstruction can result in poor reconstruction quality. However, most people in the community do not have sufficient computing resources to support the joint training of VQAE and UNet models. Then, we raise another question: \textit{Is it possible to perform an approximate joint optimization of these two models?}

To address these challenges, we propose a novel diffusion inversion method in this work, termed Latent Bias Inversion (LBI). To address the first challenge, as shown in Figure~\ref{fig: LBO}, this paper proposes the Latent Bias Optimization (LBO) strategy that learns the latent bias vector between adjacent inversion steps to align the inversion and generation trajectories, thereby achieving accurate inversion. The proposed LBO supports gradient-based, numerical, and hybrid optimization. Regarding the second challenge, we notice that the image latent representation $z_0$ obtained by the encoder of VQAE serves as the interface connecting the diffusion inversion process with VQAE reconstruction. Therefore, we perform an approximate joint optimization of these two processes by learning an improved image latent representation $z_0$, which yields significant improvements with minimal effort. We conduct extensive experiments to evaluate the proposed LBI method on image reconstruction. The comprehensive empirical results demonstrate that the proposed LBI method significantly outperforms existing state-of-the-art methods. Besides, the successful application of our method in two relevant downstream tasks (image editing and rare concept generation) further demonstrates its effectiveness.

\section{Related Work and Unique Contributions}

In this section, we first review existing diffusion inversion methods related to our work. Then, we review two relevant downstream tasks. Finally, we summarize the major contributions of this work.

\subsection{Diffusion Inversion Methods}

\begin{figure}[t]
\centering
\includegraphics[width=\columnwidth]{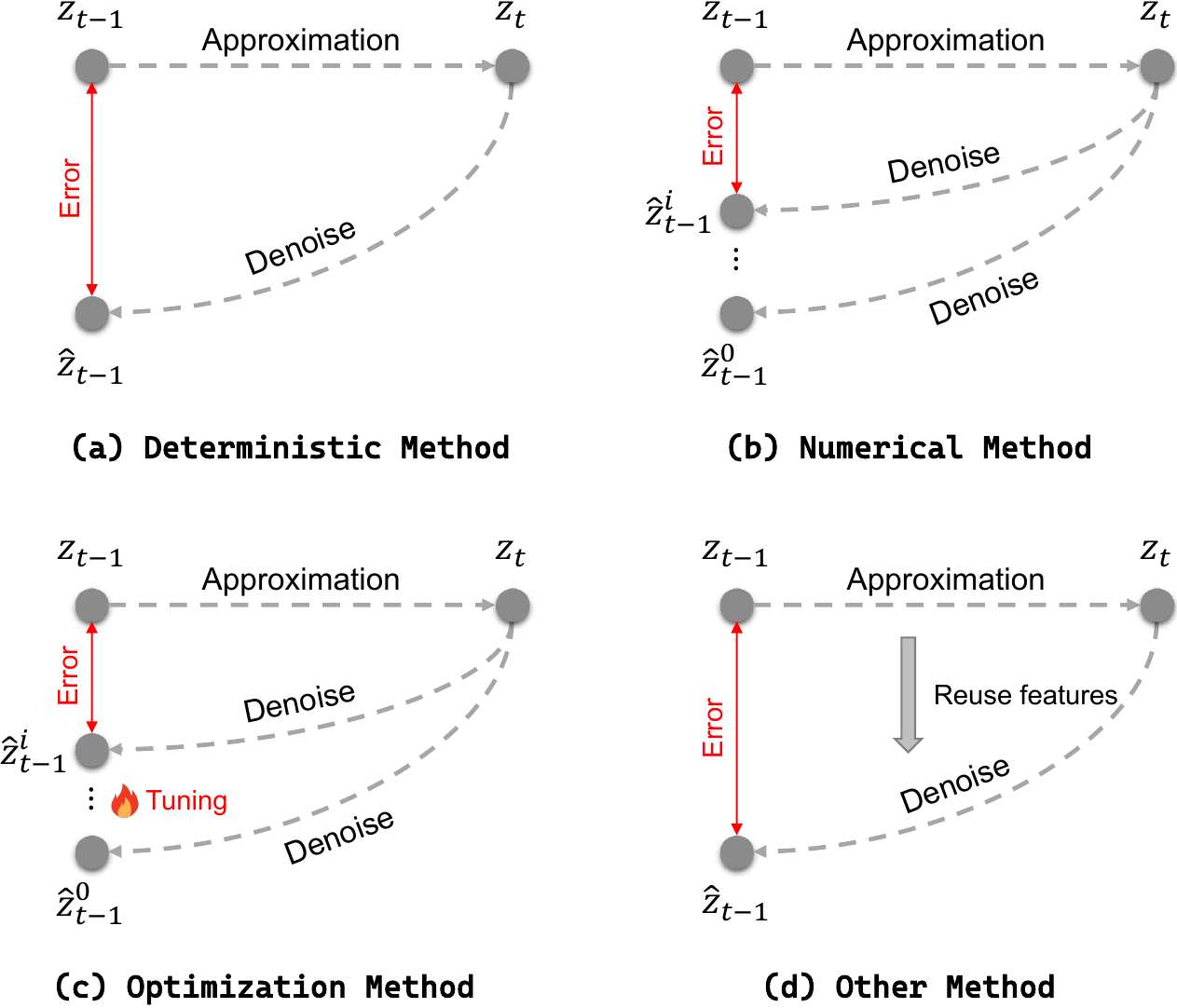}
\caption{A comparison of four different types of diffusion inversion methods.}
\label{fig: inversion methods}
\end{figure}

Since inversion becomes a fundamental building block in various tasks, several inversion methods have been proposed. As shown in Figure~\ref{fig: inversion methods}, existing diffusion inversion methods can be categorized into 4 types: deterministic, numerical, optimization-based, and other methods. 

\subsubsection{Deterministic Methods}

DDIM inversion~\cite{Song2021DDIM,Dhariwal2021DDIM} and Negative-Prompt Inversion (NPI)~\cite{Miyake2023NPI} are deterministic methods that use the deterministic DDIM sampler based on the reversible assumption of ODE processes. The only difference between DDIM inversion and NPI is that NPI omits the Classifier-Free Guidance (CFG)~\cite{Ho2021CFG} in the denoising process to achieve a relatively more precise reconstruction.

\subsubsection{Numerical Methods}

Numerical methods leverage historical states to provide more accurate approximations, which include Exact Diffusion Inversion via Coupled Transformations (EDICT)~\cite{Wallace2023EDICT}, Accelerated Iterative Diffusion Inversion (AIDI)~\cite{Pan2023AIDI}, Fixed-Point Inversion (FPI)~\cite{Meiri2023FPI}, Bi-Directional Integration Approximation (BDIA)~\cite{Zhang2024BDIA}, ReNoise~\cite{Garibi2024ReNoise}, Bidirectional Explicit Linear Multi-step sampler (BELM)~\cite{Wang2024BELM}, and Guided Newton-Raphson Inversion (GNRI)~\cite{Samuel2025GNRI}. EDICT proposes an auxiliary diffusion branch to compute both backward and forward passes and achieves exact inversion. AIDI, FPI, and GNRI are based on the numerical scheme of the fixed-point iteration technique~\cite{Burden2015NumericalAnalysis}. BDIA estimates the next diffusion state as a linear combination of the estimated Gaussian noise at the current step and the previous and current diffusion states, thereby achieving exact inversion. ReNoise refines the approximation of latent prediction by averaging multiple predictions along the inversion trajectory. BELM introduces a generic formula for the general exact inversion samples that seeks to establish a linear relation among the previous, current, and next states.

\subsubsection{Optimization Methods}

Optimization methods perform accurate inversion by training some variables. Representative optimization-based methods include Null-Text Inversion (NTI)~\cite{Mokady2023NTI}, Prompt-Tuning Inversion (PTI)~\cite{Dong2023PTI}, and Exact Inversion of DPM (ExactDPM)~\cite{Hong2024ExactDPM}. NTI optimizes the unconditional prompt embedding in each inference step to align the denoising and inversion processes. In contrast, PTI finetunes the conditional prompt embedding and achieves a more stable inversion performance. ExactDPM directly trains the target state to align the current state and its prediction.

\subsubsection{Other Methods}

Some methods reuse the features from the inversion process to reduce the error between the inversion and denoising processes. Representative works include Real-world Image Variation by ALignment (RIVAL)~\cite{Zhang2023RIVAL} and Tuning-free Inversion-enhanced Control (TIC)~\cite{Duan2024TIC}. RIVAL designs a cross-image self-attention injection for feature interaction and a step-wise distribution normalization to align the latent features. TIC achieves exact inversion by reusing the attention features of key and value in the inversion process.

Unlike previous works, in this work, we directly model the bias latent representation between adjacent steps to seek an optimal inversion trajectory for the given image, thereby achieving the exact inversion without introducing burdensome additional parameters. Additionally, our method supports both gradient descent and numerical iteration to model the biased latent representation, enabling a trade-off between accuracy and computational efficiency.

\subsection{Relevant Downstream Tasks}

Inversion establishes the connection between real-world images and diffusion models, which derive various downstream tasks. In this paper, we review two relevant downstream tasks that have been actively studied in recent literature. 

\subsubsection{Image Editing}

Recent research has shown increasing interest in leveraging the powerful generative capabilities of text-to-image diffusion models for image editing tasks. Prompt-to-Prompt (P2P)~\cite{Hertz2023P2P} injects attention maps from the inversion process guided by the source prompt into those guided by the target prompt, in order to preserve the spatial layout and geometric structure of the original image. However, P2P fails to edit real-world images due to the low reconstruction quality of DDIM inversion. MasaCtrl~\cite{Cao2023MasaCtrl} introduces a mask-guided mutual self-attention mechanism, which replaces the key and value attention features in self-attention layers to enhance the consistency of the edited image. Plug-and-Play (PnP)~\cite{Tumanyan2023PnP} enables fine-grained control over generated structures by manipulating spatial and self-attention features, directly injecting features from a guidance image.

\subsubsection{Rare Concept Generation}

Text-to-image diffusion models can synthesize high-quality images, but the long-tail nature of their training data limits their ability to fully understand concepts represented at the tail end of the distribution. The imbalance of training samples often leads to the incorrect generation of images for rare concepts. To address this challenge, SeedSelect \cite{Samuel2024SeedSelect} utilizes a small set of rare concept images as references and employs a diffusion inversion module to enhance the initial seeds iteratively. NAO-SeedSelect~\cite{Samuel2023NAOSeedSelect} further improves this process by introducing alternative paths and centroids for seed initialization. Our proposed method seeks an optimal inversion trajectory for the given reference image, thereby enabling the rare concept generation model to achieve better concept alignment in the generation process.

\subsection{Unique Contributions}

Compared to existing methods, our unique contributions include: (1) We tackle the challenge of misalignment between inversion and generation processes by learning a latent bias vector between adjacent inversion steps to minimize the misalignment errors. (2) We address the challenge of mismatches between the diffusion inversion process and VQAE reconstruction by performing an approximate joint optimization of these two processes, \textit{i.e.}, learning to adjust the image latent representation $z_0$, which serves as the connecting interface. (3) We evaluate our proposed LBI method on the entire COCO~\cite{Lin2014COCO} validation set for the image reconstruction task, demonstrating its significant performance improvement over existing methods. Furthermore, we apply the proposed method to relevant downstream tasks (\textit{e.g.}, image editing and rare concept generation), showcasing its significant potential in these downstream tasks.

\section{Proposed Method}

In this section, we first provide a brief overview of the relevant background knowledge and the proposed method. Next, we present a detailed description of our method. Finally, we demonstrate how our method is applied to two downstream tasks (image editing and rare concept generation).

\subsection{Background and Method Overview}

We present the relevant background knowledge and an overview of the method in this subsection.

\subsubsection{Background}

Latent Diffusion Models (LDMs)~\cite{Rombach2022LDM} perform the diffusion process in a compressed latent space rather than in the pixel space, thereby significantly improving computational efficiency. LDMs use a pre-trained encoder $\mathcal{E}$ to project the input image $x_0$ into a latent space, from which the image can be reconstructed using a corresponding pre-trained decoder $\mathcal{D}$, \textit{i.e.,}
\begin{equation}
\hat{x}_0 = \mathcal{D}\left ( \mathcal{E} \left ( x_0 \right) \right) .
\label{eq: vqae}
\end{equation}
Here, the VQAE reconstruction can be regarded as the upper bound of inversion. Following the DDPM scheme, the forward pass gradually adds noise to the clean image latent representation $z_0$ until it becomes standard Gaussian noise:
\begin{equation}
z_t = \sqrt{1 - \beta_t} z_{t-1} + \sqrt{\beta_t} \epsilon_t, \quad t=1,2,\dots,T,
\end{equation}
where $\beta_t$ is the parameter at timestep $t$ predefined by the scheduler, $\epsilon_t \sim \mathcal{N} (0, \mathbf{I})$ is the random sampled standard Gaussian noise. Then the forward pass is equivalent to:
\begin{equation}
z_t = \sqrt{\bar{\alpha}_t} z_0 + \sqrt{1 - \bar{\alpha}_t} \epsilon_t,
\end{equation}
where $\bar{\alpha}_t = {\textstyle \prod_{s=1}^{t} (1-\beta_s)}$. The backward pass progressively generates an image from an initial noise sample $z_T$ guided by a textual condition:
\begin{equation}
z_{t-1} = \phi_t z_t + \psi_t \mathbf{F}_\theta (z_t, t, \mathcal{C}) + \sigma_t \epsilon_t,
\end{equation}
where $\mathbf{F}_\theta$ denotes a pre-trained diffusion model and $\mathcal{C} = \tau_\theta \left ( \mathcal{P} \right )$ represents the embedding of the conditional text prompt obtained by a pre-trained CLIP~\cite{CLIP} text encoder $\tau_\theta$, and
\begin{equation}
\begin{array}{c}
\phi_t = \sqrt{\frac{\bar{\alpha}_{t-1}}{\bar{\alpha}_t}}, \quad \sigma_t = \eta \sqrt{\frac{\beta_t (1 - \bar{\alpha}_{t-1})}{1 - \bar{\alpha}_t}}, \\
\psi_t = \sqrt{1 - \bar{\alpha}_{t-1} - \sigma_{t}^{2}} - \sqrt{\frac{(1 - \bar{\alpha}_t) \bar{\alpha}_{t-1}}{\bar{\alpha}_t}}.
\end{array}
\end{equation}
Here, the coefficient $\eta \in [0, 1]$ controls the stochasticity of the sampling process. When $\eta = 1$, the process corresponds to the original DDPM sampling, while $\eta = 0$ corresponds to the deterministic DDIM sampler. In this work, we focus on the deterministic case with $\eta = 0$:
\begin{equation}
\frac{z_{t-1}}{\sqrt{\bar{\alpha}_{t-1}}} = \frac{z_{t}}{\sqrt{\bar{\alpha}_{t}}} + \left ( \sqrt{\frac{1}{\bar{\alpha}_{t-1}} - 1} - \sqrt{\frac{1}{\bar{\alpha}_t} - 1} \right ) \mathbf{F}_\theta (z_t, t, \mathcal{C}).
\label{eq: ddim denoise}
\end{equation}
Since the sampling process in diffusion models can alternatively be viewed as solving the corresponding ODEs~\cite{Lu2022DPMSolver}, the denoising process can be reversed under the assumption that the ODE process is reversible in the limit of small steps. Formally, the inversion equation can be expressed as:
\begin{equation}
\frac{z_{t}}{\sqrt{\bar{\alpha}_{t}}} = \frac{z_{t-1}}{\sqrt{\bar{\alpha}_{t-1}}} + \left ( \sqrt{\frac{1}{\bar{\alpha}_{t}} - 1} - \sqrt{\frac{1}{\bar{\alpha}_{t-1}} - 1} \right ) \mathbf{F}_\theta (z_{t-1}, t, \mathcal{C}).
\label{eq: ddim inversion}
\end{equation}
Equation (\ref{eq: ddim inversion}) also corresponds to the well-known DDIM Inversion method~\cite{Song2021DDIM,Dhariwal2021DDIM}.
The approximation error introduced by this assumption leads to the misalignment between inversion and sampling processes, which is the fundamental problem in diffusion inversion methods.

\subsubsection{Method Overview}

We identify two major challenges that limit the accurate inversion of real-world images using diffusion models: the misalignment between the inversion and generation trajectories, and the mismatch between the diffusion inversion process and VQAE reconstruction. We provide feasible solutions to address these two challenges. We introduce the Latent Bias Optimization (LBO) strategy to align the inversion and generation trajectories by optimizing a latent bias vector at each inversion step. The proposed LBO supports gradient-based, numerical, and hybrid optimization, and all three variants exhibit robust performance. Additionally, we propose the Image Latent Boosting (ILB) technique to perform an approximate joint optimization of the diffusion inversion process and VQAE reconstruction by adjusting the image latent representation $z_0$, which serves as the connecting interface of these two processes. Together with the LBO strategy and ILB technique, we propose the Latent Bias Inversion (LBI) method that achieves accurate and stable inversions for diffusion models. In addition, we successfully apply the proposed method to various downstream tasks, acting as a powerful seed noise generator for real-world images.

\subsection{Latent Bias Optimization}

From Equations (\ref{eq: ddim denoise}) and (\ref{eq: ddim inversion}), it follows that the core assumption underlying DDIM inversion is
\begin{equation}
\mathbf{F}_\theta (z_t, t, \mathcal{C}) \approx \mathbf{F}_\theta (z_{t-1}, t, \mathcal{C}).
\end{equation}
In other words, it assumes $z_t \approx z_{t-1}$ for the implicit calculation of the UNet model. However, as the errors introduced by this assumption accumulate, the denoising trajectory progressively diverges from the inversion trajectory, ultimately degrading image reconstruction performance. To address this critical issue, we directly optimize the latent bias between adjacent inversion steps to align the inversion and generation trajectories. Specifically, for a single inversion step, we first introduce a latent bias vector with the same dimension as the latent vector in LDMs to redefine the inversion and generation processes:
\begin{equation}
z_t = z_{t-1} + b_t ,
\label{eq: LBO inverse}
\end{equation}
\begin{equation}
z_{t-1} = z_t - \tilde{b}_t .
\label{eq: LBO denoise}
\end{equation}
Here, to distinguish the inversion and generation processes, we denote the latent bias vector in the inversion process as $b_t$, and that in the generation process as $\tilde{b}_t$:
\begin{equation}
\tilde{b}_t = \left ( 1 - \phi_t \right ) z_t - \psi_t \mathbf{F}_\theta \left ( z_t, t, \mathcal{C} \right ) .
\label{eq: bt denoise}
\end{equation}
Although the ODE approximation cannot yield an accurate inversion result, it provides a reasonable estimation of the next latent state $z_t$. Therefore, the latent bias vector in the inversion process can be initialized by:
\begin{equation}
b_t^0 = \frac{1-\phi_{t}}{\phi_{t}} z_{t-1} - \frac{\psi_{t}}{\phi_{t}} \mathbf{F}_\theta \left ( z_{t-1}, t, \mathcal{C} \right ) .
\end{equation}
Note that the latent bias vector is an auxiliary variable and does not need to be retained for subsequent use. Consequently, the proposed LBO strategy does not incur significant memory overhead. Our proposed LBO algorithm is presented in Algorithm~\ref{alg: LBO}. Here, we suggest three methods to optimize the latent bias vector at each step, and all methods yield accurate and robust inversion results.

\begin{algorithm}[t]
\caption{Latent Bias Optimization}
\label{alg: LBO}

\textbf{Input}: The image latent representation $z_0$ and the corresponding textual prompt $\mathcal{P}$\\
\textbf{Output}: The corresponding initial noise sample $z_T$

\begin{algorithmic}[1] 
\STATE $\mathcal{C} = \tau_\theta \left ( \mathcal{P} \right )$
\FOR{$t=1,2,\cdots,T$}
\STATE $i \leftarrow 0$
\IF{gradient-based optimization}
\REPEAT 
\STATE $b_t^{i+1} \leftarrow b_t^i - \eta \nabla_{b_t} | b_t^i - \tilde{b}_t^i | $
\UNTIL{converged}
\ELSIF{numerical optimization}
\REPEAT
\STATE $b_{t}^{i+1} \leftarrow (1 - \phi_t) (z_{t-1} + b_t^i) + \psi_t \mathbf{F}_\theta (z_{t-1}+b_t^i, t, \mathcal{C})$
\UNTIL{converged}
\ELSIF{hybrid optimization}
\STATE cnt $\leftarrow$ 0
\REPEAT
\IF{cnt $<N_g$}
\STATE $b_t^{i+1} \leftarrow b_t^i - \eta \nabla_{b_t} | b_t^i - \tilde{b}_t^i | $
\ELSE
\STATE $b_{t}^{i+1} \leftarrow (1 - \phi_t) (z_{t-1} + b_t^i) + \psi_t \mathbf{F}_\theta (z_{t-1}+b_t^i, t, \mathcal{C})$
\ENDIF
\STATE cnt += 1
\UNTIL{converged}
\ENDIF
\STATE $z_t = z_{t-1} + b_t$
\ENDFOR
\STATE \textbf{return} $z_T$
\end{algorithmic}
\end{algorithm}

\subsubsection{Gradient-based Optimization}

A straightforward method is to learn the latent bias vector $b_t$ by solving the following objective using gradient descent:
\begin{equation}
b_t = \arg \min_{b_t} | b_t - \tilde{b}_t | ,
\label{eq: LBO gradient}
\end{equation}
which aligns the inversion and generation trajectories while keeping the parameters of the diffusion UNet model frozen.

\subsubsection{Numerical Optimization}

However, the backpropagation operation in the gradient-based method often requires substantial computational time, which limits its practicality in time-sensitive applications. To improve efficiency, we employ a numerical method that is approximately 10× faster than the gradient-based method while maintaining comparable performance. Specifically, the latent bias vector in each inversion step is updated following:
\begin{equation}
b_{t}^{i+1} = (1 - \phi_t) (z_{t-1} + b_t^i) + \psi_t \mathbf{F}_\theta (z_{t-1}+b_t^i, t, \mathcal{C}) .
\end{equation}

\subsubsection{Hybrid Optimization}

Although the numerical method is computationally efficient, its convergence often relies on the quality of the initial estimation. Therefore, we develop a hybrid strategy that optimizes the latent bias vector using the gradient-based method for the first $N_g$ steps to obtain a better initial estimation, and then switches to the numerical method to accelerate the subsequent process.

\subsection{Image Latent Boosting}

\begin{algorithm}[t]
\caption{Image Latent Boosting}
\label{alg: ILB}

\textbf{Input}: The input image $x_0$, the timestep interval $\delta_t$, and the corresponding textual prompt $\mathcal{P}$\\
\textbf{Output}: The optimized image latent representation $\hat{z}_0$

\begin{algorithmic}[1] 
\STATE $\mathcal{C} = \tau_\theta \left ( \mathcal{P} \right )$
\STATE $\hat{z}_0 \leftarrow \mathcal{E} (x_0)$
\REPEAT
\STATE $\hat{x}_0 \leftarrow \mathcal{D} (\hat{z}_0)$
\STATE $\mathcal{L}_{\text{con}} \leftarrow |x_0 - \hat{x}_0| - \text{SSIM} (x_0, \hat{x}_0) + \text{LPIPS} (x_0, \hat{x}_0)$
\STATE $\hat{z}_{\delta_t} \leftarrow \frac{1}{\phi_{0,\delta_t}} \hat{z}_0 - \frac{\psi_{0,\delta_t}}{\phi_{0,\delta_t}} \mathbf{F}_\theta (\hat{z}_0, \delta_t, \mathcal{C})$
\STATE $\hat{z}_{0 | \delta_t} \leftarrow \phi_{0,\delta_t} \hat{z}_{\delta_t} + \psi_{0,\delta_t} \mathbf{F}_\theta (\hat{z}_{\delta_t}, \delta_t, \mathcal{C})$
\STATE $\mathcal{L}_{\text{reg}} \leftarrow | \hat{z}_0 - \hat{z}_{0 | \delta_t} |$
\STATE $\hat{z}_0 \leftarrow \hat{z}_0 - \gamma \nabla_{\hat{z}_0} (\mathcal{L}_{\text{con}} + \mathcal{L}_{\text{reg}})$
\UNTIL{converged}
\end{algorithmic}
\end{algorithm}

\begin{figure}
\centering
\includegraphics[width=\linewidth]{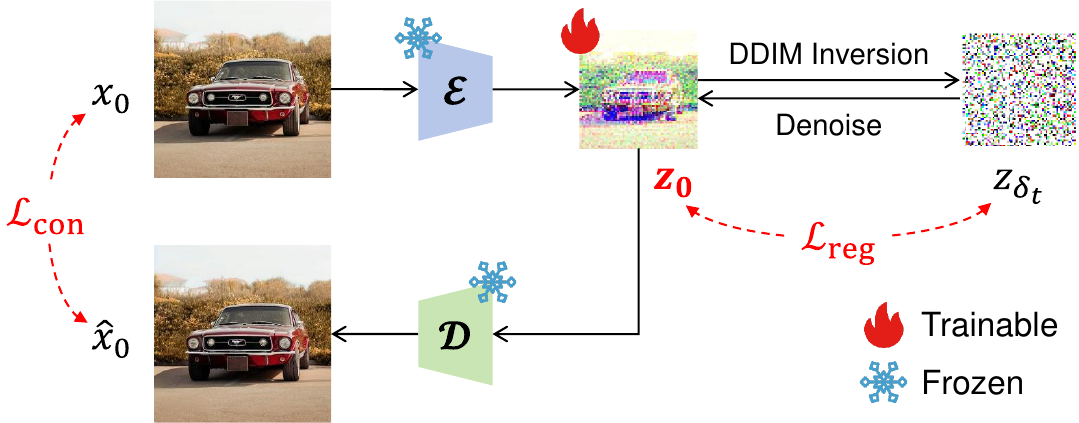}
\caption{The overview of the proposed Image Latent Boosting (ILB) technique.}
\label{fig: ILB}
\end{figure}

The mismatch between the diffusion process and VQAE reconstruction is also a serious problem that limits the perfect reconstruction of real-world images. We observe that the reconstruction Peak Signal-to-Noise Ratio (PSNR) of the VQAE in Stable Diffusion (SD) v1.5 ranges from 25 to 31 dB for general images. The reconstruction performance would be even worse for some images with rich textures or textual content. However, retraining the VQAE is non-trivial and may introduce misalignment with the pre-trained diffusion UNet model. We observe that the image latent representation $z_0 = \mathcal{E} (x_0)$ serves as the connecting interface between the VQAE and the diffusion UNet model. Based on this observation, as shown in Figure~\ref{fig: ILB}, we propose the Image Latent Boosting (ILB) method, which optimizes the $z_0$ while keeping the parameters of VQAE and diffusion UNet frozen, yielding significant improvements with minimal effort. Specifically, we design a loss function that enforces consistency between the input image $x_0$ and its reconstruction $\hat{x}_0$, while simultaneously ensuring alignment with the diffusion UNet model. Regarding the color, texture, and structure consistency, we use the combination of L1 distance, Structural Similarity (SSIM), and Learned Perceptual Image Patch Similarity (LPIPS)~\cite{Zhang2018LPIPS} perceptual similarity to be the consistency loss, which is defined as:
\begin{equation}
\begin{aligned}
\mathcal{L}_{\text{con}} (x_0, z_0) = & | x_0 - \mathcal{D}(z_0) | - \text{SSIM}(x_0, \mathcal{D}(z_0)) \\
& + \text{LPIPS} (x_0, \mathcal{D}(z_0)) .
\end{aligned}
\label{eq: consistency loss}
\end{equation}
Because the VQAE and diffusion UNet model are trained separately, the alignment of these two models is also important. To ensure the alignment between the optimized $z_0$ and the diffusion UNet model, we introduce an additional regularization term:
\begin{equation}
\mathcal{L}_{\rm{reg}} (z_0, \delta_t) = | z_0 - z_{0 | \delta_t} | ,
\label{eq: regularization loss}
\end{equation}
where $\delta_t$ denotes the timestep interval, and $z_{0 | \delta_t}$ represents the reconstructed image latent obtained from DDIM inversion:
\begin{equation}
z_{0 | \delta_t} = \phi_{0,\delta_t} z_{\delta_t} + \psi_{0,\delta_t} \mathbf{F}_\theta (z_{\delta_t}, \delta_t, \mathcal{C}),
\end{equation}
\begin{equation}
z_{\delta_t} = \frac{1}{\phi_{0,\delta_t}} z_0 - \frac{\psi_{0,\delta_t}}{\phi_{0,\delta_t}} \mathbf{F}_\theta (z_0, \delta_t, \mathcal{C}) .
\end{equation}
Here,
\begin{equation}
\phi_{0,\delta_t} = \sqrt{\frac{\bar{\alpha}_0}{\bar{\alpha}_{\delta_t}}}, \quad \psi_{0,\delta_t} = \sqrt{1-\bar{\alpha}_0} - \sqrt{\frac{(1-\bar{\alpha}_{\delta_t})\bar{\alpha}_0}{\bar{\alpha}_{\delta_t}}}.
\end{equation}
The ILB loss is defined as the combination of the consistency loss and the regularization loss. We optimize $z_0$ with this loss using gradient descent to obtain an improved image latent representation. The proposed ILB algorithm is summarized in Algorithm~\ref{alg: ILB}.

\subsection{Application to Downstream Tasks}

Inversion bridges real-world images and generative models, thereby enabling various downstream tasks. We believe that a more accurate inversion method, capable of faithfully reconstructing the given image, will enhance the consistency with the original image in downstream tasks. In this subsection, we apply the proposed LBI method to various downstream tasks, where it serves as a powerful initial noise generator for real-world images.

\subsubsection{Image Editing}
Image editing is a popular research area, and many methods have utilized diffusion models for different types of image editing. A common approach among these methods requires inversion to edit real images. This involves obtaining a seed noise $z_T$ capable of reconstructing the original image via the denoising process of a pre-trained diffusion model. Specifically, in this method, two denoising processes are carried out simultaneously using $z_T$. One process uses the source prompt to reconstruct the image, while the other injects features from the first process to preserve some attributes of the original image while manipulating other aspects. Therefore, the quality of the seed noise sample directly affects the consistency between the edited image and the original image. The proposed LBI method can be integrated into existing image editing pipelines to enhance the faithfulness of the edited results to the original image.

\subsubsection{Rare Concept Generation}

Our method is a flexible plug-and-play module, whose effectiveness can be well validated through the rare concept generation task. Specifically, we integrate the proposed LBI method into NAO-SeedSelect~\cite{Samuel2023NAOSeedSelect} by replacing only the VAE encoder module to improve natural appearance consistency learning. Compared with the original VAE encoder, our method produces more informative latent representations, which provide more reliable guidance for optimizing the initial generation point during the back-propagation process of the denoising model. As a result, higher-quality images of rare concepts are generated.

\section{Experimental Results}

In this section, we conduct extensive experiments to evaluate the effectiveness of the proposed LBI method. Specifically, we first apply it to the image reconstruction task to validate its effectiveness. In addition, we further evaluate LBI on two representative downstream tasks to demonstrate its potential in broader applications.

\begin{table}[htbp]
\centering
\caption{Comparison of different methods across three variants of diffusion models on the COCO 2017 validation set.}
\label{tab: comparisons image reconstruction}
\resizebox{\linewidth}{!}{
\begin{tabular}{lcccc}
\toprule
\multirow{2}{*}{\textbf{Method}} & \multirow{2}{*}{\textbf{PSNR ($\uparrow$)}} & \multirow{2}{*}{\textbf{SSIM ($\uparrow$)}} & \multirow{2}{*}{\textbf{LPIPS ($\downarrow$)}} & \textbf{Additional} \\
 & & & & \textbf{Parameters} \\
\midrule
\multicolumn{5}{l}{\textbf{\textit{Stable Diffusion v1.5 (SD1)}}} \\
VQAE (upper bound) & 26.75 & 0.7699 & 0.0512 & 0 \\
DDIM Inversion & 14.10 & 0.4480 & 0.5210 & 0 \\
NPI & 24.00 & 0.7208 & 0.1141 & 0 \\
NTI & 25.81 & 0.7566 & 0.0765 & 2.81 M \\
EDICT & 26.18 & 0.7581 & 0.0723 & 16 K \\
AIDI-A & 26.38 & 0.7595 & 0.0624 & 0 \\
AIDI-E & 25.98 & 0.7565 & 0.0803 & 0 \\
PTI & 26.11 & 0.7581 & 0.0747 & 2.81 M \\
FPI & 26.59 & 0.7671 & 0.0550 & 0 \\
RIVAL & 21.68 & 0.7042 & 0.1193 & 1.08 G \\
TIC & 26.69 & 0.7684 & 0.0546 & 759.38 M \\
ReNoise & 26.12 & 0.7593 & 0.0660 & 16 K \\
BDIA & 24.58 & 0.7387 & 0.0887 & 16 K \\
BELM & 26.71 & 0.7688 & 0.0518 & 16 K \\
GNRI & 24.05 & 0.7232 & 0.1139 & 0 \\
ExactDPM (w/o DI) & 26.27 & 0.7603 & 0.0667 & 0 \\
ExactDPM & 27.29 & 0.7791 & 0.1989 & 0 \\
\ours{LBI-G (w/o ILB)} & \ours{26.63} & \ours{0.7691} & \ours{0.0557} & \ours{0} \\
\ours{LBI-N (w/o ILB)} & \ours{26.65} & \ours{0.7680} & \ours{0.0547} & \ours{0} \\
\ours{LBI-H (w/o ILB)} & \ours{26.54} & \ours{0.7663} & \ours{0.0560} & \ours{0} \\
\ours{LBI-G} & \ours{\textbf{28.14}} & \ours{\textbf{0.8338}} & \ours{\textbf{0.0332}} & \ours{0} \\
\ours{LBI-N} & \ours{\underline{28.12}} & \ours{\underline{0.8328}} & \ours{\underline{0.0344}} & \ours{0} \\
\ours{LBI-H} & \ours{28.04} & \ours{0.8310} & \ours{0.0372} & \ours{0} \\

\midrule
\multicolumn{5}{l}{\textbf{\textit{Stable Diffusion v2.1 (SD2)}}} \\
VQAE (upper bound) & 30.09 & 0.8368 & 0.0423 & 0 \\
DDIM Inversion & 14.80 & 0.4997 & 0.4533 & 0 \\
NPI & 21.19 & 0.6638 & 0.2582 & 0 \\
NTI & 23.67 & 0.7349 & 0.1635 & 3.76 M \\
EDICT & 27.68 & 0.8053 & 0.0927 & 36 K \\
AIDI-A & 26.85 & 0.7765 & 0.1446 & 0 \\
AIDI-E & 26.87 & 0.7770 & 0.1441 & 0 \\
PTI & 27.50 & 0.8059 & 0.0768 & 3.76 M \\
FPI & 26.83 & 0.7765 & 0.1447 & 0 \\
RIVAL & 27.39 & 0.8131 & 0.0612 & 2.42 G \\
TIC & 27.51 & 0.8112 & 0.0591 & 1.67 G \\
ReNoise & 27.48 & 0.8094 & 0.0821 & 36 K \\
BDIA & 26.65 & 0.8035 & 0.0898 & 36 K \\
BELM & 19.02 & 0.6786 & 0.2074 & 36 K \\
GNRI & 21.23 & 0.6671 & 0.2571 & 0 \\
ExactDPM (w/o DI) & 28.56 & 0.8242 & 0.0551 & 0 \\
ExactDPM & 30.08 & 0.8504 & 0.0942 & 0 \\
\ours{LBI-G (w/o ILB)} & \ours{29.62} & \ours{0.8320} & \ours{0.0497} & \ours{0} \\
\ours{LBI-N (w/o ILB)} & \ours{29.57} & \ours{0.8307} & \ours{0.0508} & \ours{0} \\
\ours{LBI-H (w/o ILB)} & \ours{29.65} & \ours{0.8318} & \ours{0.0493} & \ours{0} \\
\ours{LBI-G} & \ours{\underline{30.65}} & \ours{\underline{0.8869}} & \ours{\underline{0.0336}} & \ours{0} \\
\ours{LBI-N} & \ours{30.57} & \ours{0.8850} & \ours{0.0353} & \ours{0} \\
\ours{LBI-H} & \ours{\textbf{30.72}} & \ours{\textbf{0.8872}} & \ours{\textbf{0.0327}} & \ours{0} \\

\midrule
\multicolumn{5}{l}{\textbf{\textit{Stable Diffusion XL v1.0 (SDXL)}}} \\
VQAE (upper bound) & 32.28 & 0.8834 & 0.0407 & 0 \\
DDIM Inversion & 16.04 & 0.6093 & 0.3936 & 0 \\
NPI & 25.93 & 0.8065 & 0.1542 & 0 \\
NTI & 26.61 & 0.8340 & 0.1077 & 7.58 M \\
EDICT & 31.84 & 0.8757 & 0.0482 & 64 K \\
AIDI-A & 31.71 & 0.8784 & 0.0475 & 0 \\
AIDI-E & 31.67 & 0.8785 & 0.0477 & 0 \\
PTI & 30.11 & 0.8629 & 0.0695 & 7.58 M \\
FPI & 31.60 & 0.8777 & 0.0480 & 0 \\
RIVAL & 20.04 & 0.7446 & 0.1777 & 9.77 G \\
TIC & 31.47 & 0.8678 & 0.0622 & 10.88 G \\
ReNoise & 29.76 & 0.8561 & 0.0805 & 64 K \\
BDIA & 26.96 & 0.8320 & 0.0936 & 64 K \\
BELM & 31.90 & 0.8773 & 0.0437 & 64 K \\
GNRI & 26.05 & 0.8129 & 0.1501 & 0 \\
ExactDPM (w/o DI) & 28.28 & 0.8187 & 0.1368 & 0 \\
ExactDPM & 30.33 & 0.8563 & 0.1281 & 0 \\
\ours{LBI-G (w/o ILB)} & \ours{31.58} & \ours{0.8782} & \ours{0.0485} & \ours{0} \\
\ours{LBI-N (w/o ILB)} & \ours{31.48} & \ours{0.8762} & \ours{0.0498} & \ours{0} \\
\ours{LBI-H (w/o ILB)} & \ours{31.61} & \ours{0.8777} & \ours{0.0482} & \ours{0} \\
\ours{LBI-G} & \ours{\textbf{32.52}} & \ours{\textbf{0.9211}} & \ours{\textbf{0.0369}} & \ours{0} \\
\ours{LBI-N} & \ours{32.41} & \ours{0.9203} & \ours{\underline{0.0373}} & \ours{0} \\
\ours{LBI-H} & \ours{\underline{32.48}} & \ours{\underline{0.9207}} & \ours{0.0377} & \ours{0} \\
\bottomrule
\end{tabular}
}
\end{table}

\subsection{Experimental Setup}

We evaluate the proposed method on the image reconstruction task and the other two downstream tasks, \textit{i.e.}, image editing and rare concept generation.

\subsubsection{Baselines}

For the image reconstruction task, we compare the proposed LBI method with existing diffusion inversion methods, including DDIM Inversion~\cite{Song2021DDIM,Dhariwal2021DDIM}, NPI~\cite{Miyake2023NPI}, NTI~\cite{Mokady2023NTI}, EDICT~\cite{Wallace2023EDICT}, AIDI~\cite{Pan2023AIDI}, PTI~\cite{Dong2023PTI}, FPI~\cite{Meiri2023FPI}, RIVAL~\cite{Zhang2023RIVAL}, TIC~\cite{Duan2024TIC}, ReNoise~\cite{Garibi2024ReNoise}, BDIA~\cite{Zhang2024BDIA}, BELM~\cite{Wang2024BELM}, GNRI~\cite{Samuel2025GNRI}, and ExactDPM~\cite{Hong2024ExactDPM}, across three diffusion model variants: Stable Diffusion v1.5 (SD1), Stable Diffusion v2.1 (SD2), and Stable Diffusion XL v1.0 (SDXL). For the image editing task, we integrate the inversion methods into an existing editing pipeline (P2P~\cite{Hertz2023P2P}) to enable consistent editing on real-world images. To ensure a fair and comprehensive evaluation, all methods are implemented based on SD1, and we consider five representative baselines across three categories of inversion methods, \textit{i.e.}, the deterministic method (DDIM Inversion~\cite{Song2021DDIM,Dhariwal2021DDIM}), numerical methods (FPI~\cite{Meiri2023FPI} and AIDI~\cite{Pan2023AIDI}), and optimization-based methods (NTI~\cite{Mokady2023NTI} and ExactDPM~\cite{Hong2024ExactDPM}). For the rare concept generation task, we employ the proposed method in the classic rare concept generation pipeline NAO-SeedSelect~\cite{Samuel2023NAOSeedSelect}.

\subsubsection{Datasets and Evaluation Metrics}

To comprehensively evaluate the effectiveness of the proposed method, we report the image reconstruction results on the entire COCO~\cite{Lin2014COCO} 2017 validation set, using Peak Signal-to-Noise Ratio (PSNR), SSIM, and LPIPS as evaluation metrics. We evaluate the image editing performance on the PIE-Bench dataset~\cite{Ju2024PIEBench}. To quantify the alignment between the edited image and the target prompt, we report the CLIP score~\cite{Radford2021CLIP}. In addition, PSNR, SSIM, and the structural distance metric~\cite{Ju2024PIEBench} are used to assess the fidelity of the non-edited regions with respect to the original image. For the rare concept generation task, following the setup in~\cite{Samuel2024SeedSelect}, we randomly sample 20 images from rare concept categories in ImageNet-1K~\cite{deng2009imagenet} as reference images. We then measure the similarity between the reference and generated images using the CLIP image encoder.

\subsubsection{Implementation details}

The proposed method and all competing methods are implemented with the PyTorch framework. All experiments are conducted on a single NVIDIA RTX A6000 GPU. We adopt the original hyperparameter configurations of the diffusion models unless otherwise specified. The number of inference steps is set to 50 for all diffusion variants. For SD1, SD2, and SDXL, the image resolutions are set to $512 \times 512$, $768 \times 768$, and $1024 \times 1024$, respectively. We adopt the Adam optimizer for all optimization-based methods. The learning rate $\eta$ in Algorithm~\ref{alg: LBO} is set to 0.001, while the learning rate $\gamma$ in Algorithm~\ref{alg: ILB} is set to 0.1 for SD1 and SD2, and 0.01 for SDXL. For the image reconstruction task, the guidance scale of CFG is set to 1.0. For other downstream tasks, the guidance scale is configured to 7.5 for SD1 and SD2, and 5.0 for SDXL.

\begin{table}
\centering
\caption{Ablation study on the loss components of ILB.}
\label{tab: ablation ILB loss}
\resizebox{\linewidth}{!}{
\begin{tabular}{ccccccc}
\toprule
\multicolumn{3}{c}{\textbf{$\mathcal{L}_{\rm{con}}$}} & \multirow{2}{*}{\textbf{$\mathcal{L}_{\rm{reg}}$}} & \multirow{2}{*}{\textbf{PSNR ($\uparrow$)}} & \multirow{2}{*}{\textbf{SSIM ($\uparrow$)}} & \multirow{2}{*}{\textbf{LPIPS ($\downarrow$)}} \\
$\mathcal{L}_{\rm{L1}}$ & $\mathcal{L}_{\rm{SSIM}}$ & $\mathcal{L}_{\rm{LPIPS}}$ &  &  &  &  \\
\midrule
\ding{55} & \ding{55} & \ding{55} & \ding{55} & 26.65 & 0.7680 & 0.0547 \\
\ding{51} & \ding{55} & \ding{55} & \ding{55} & 29.16 & 0.8287 & 0.1055 \\
\ding{51} & \ding{51} & \ding{55} & \ding{55} & 28.44 & 0.8456 & 0.1101 \\
\ding{51} & \ding{51} & \ding{51} & \ding{55} & 28.08 & 0.8310 & 0.0312 \\
\ding{51} & \ding{51} & \ding{51} & \ding{51} & 28.12 & 0.8328 & 0.0344 \\
\bottomrule
\end{tabular}
}
\end{table}

\subsection{Comparisons on Image Reconstruction Task}

We evaluate the proposed method on the entire COCO 2017 validation set. Specifically, for each image, we randomly select one out of five captions to construct 5,000 caption-image pairs. The comparison results are reported in Table~\ref{tab: comparisons image reconstruction}, with all methods re-implemented according to their original literature to ensure a fair comparison. Notably, the VQAE method represents the upper bound of reconstruction performance, as it directly decodes the image latent representation $z_0$ encoded by the encoder. The reported results of the proposed method correspond to three variants with different optimization strategies, \textit{i.e.}, gradient-based (LBI-G), numerical (LBI-N), and hybrid (LBI-H) optimization. Based on the reported results, we make the following observations: (1) The proposed LBI method consistently outperforms all comparison methods across the three diffusion model variants without introducing additional parameters. (2) All variants of the proposed method deliver comparable performance. Thus, the numerical variant emerges as the optimal choice, as it runs considerably faster than the others. (3) When the ILB module is deactivated, all variants of the proposed LBI method consistently achieve near-perfect reconstruction performance across the three diffusion models. Notably, while BELM~\cite{Wang2024BELM} slightly outperforms the proposed method on SD1 and SDXL, LBI maintains stable performance on SD2, whereas BELM exhibits a significant performance drop. (4) The proposed ILB strategy consistently outperforms the Decoder Inversion (DI) proposed in ExactDPM~\cite{Hong2024ExactDPM} across all metrics.

\subsection{Ablation Study}

\begin{figure}
\centering
\includegraphics[width=\linewidth]{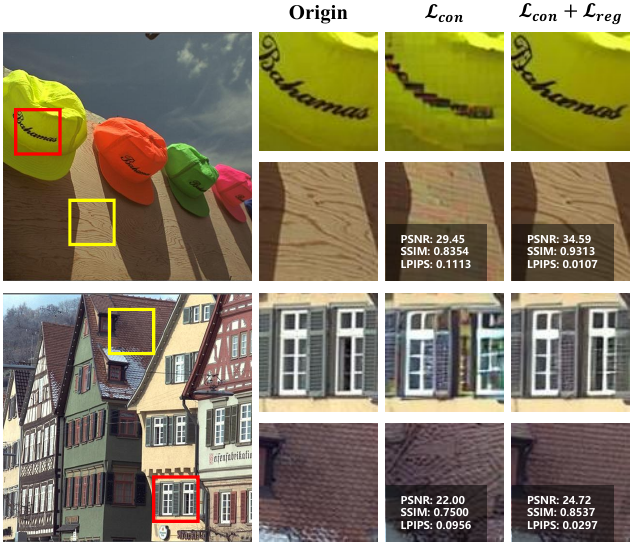}
\caption{Visual comparisons of diffusion inversion reconstruction results with and without the regularization term $\mathcal{L}_{\rm{reg}}$.}
\label{fig: ablation ILB regularization loss}
\end{figure}

\begin{figure*}
\centering
\includegraphics[width=\linewidth]{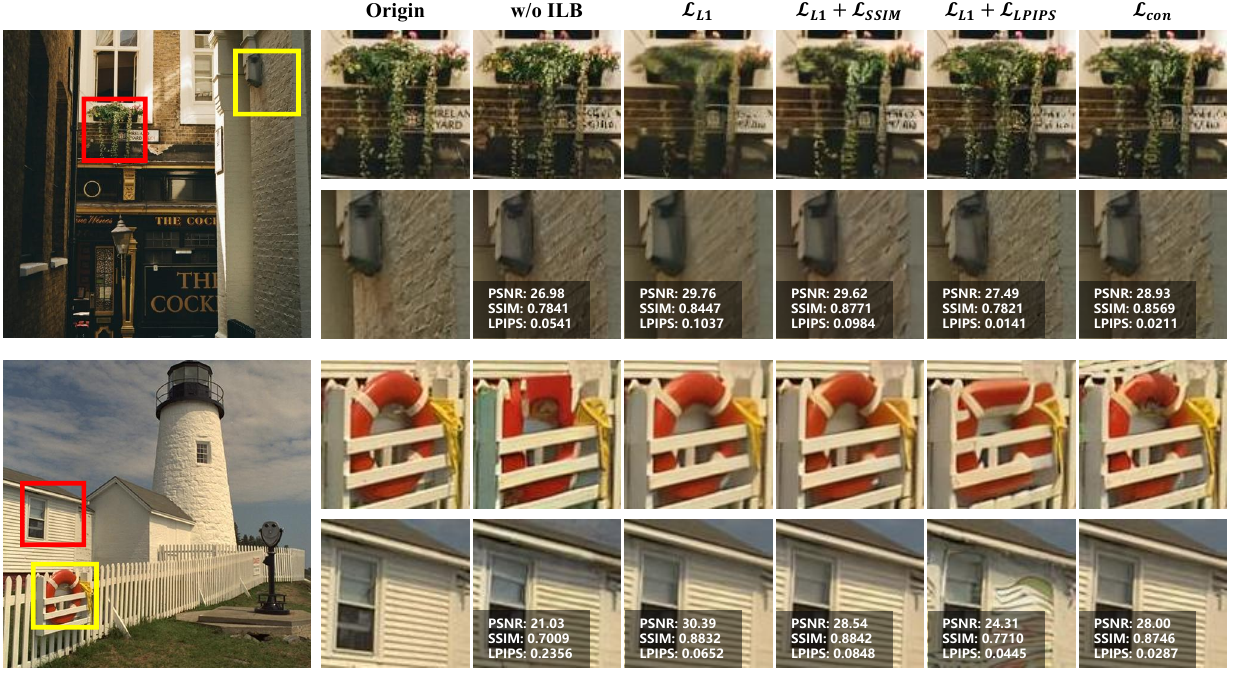}
\caption{Visualization of diffusion inversion reconstruction results under different ILB consistency loss settings.}
\label{fig: ablation ILB consistency loss}
\end{figure*}

We conduct an ablation study on the loss components of ILB to assess the contribution of each loss term. Here, we adopt SD1 as the diffusion baseline. The consistency loss is composed of an L1 loss, SSIM loss, and LPIPS perceptual loss. As shown in Table~\ref{tab: ablation ILB loss}, the L1 and SSIM losses primarily contribute to improving pixel-level consistency in the reconstructed images. However, the excessive focus on pixel-level consistency can lead to an imbalance at the perceptual level. As shown in Figure~\ref{fig: ablation ILB consistency loss}, using only the $\mathcal{L}_{\rm{L1}}$ as the consistency loss improves pixel-level consistency but causes a substantial loss of texture details, leading to overly smoothed images. Although the combination of $\mathcal{L}_{\rm{L1}}$ and $\mathcal{L}_{\rm{SSIM}}$ as the consistency loss preserves image texture details to some extent, the overall visual quality remains suboptimal. In addition, combining the $\mathcal{L}_{\rm{L1}}$ and $\mathcal{L}_{\rm{LPIPS}}$ improves the perceptual quality of images but may lead to unstable reconstruction results with severe distortions. By contrast, combining $\mathcal{L}_{\rm{L1}}$, $\mathcal{L}_{\rm{SSIM}}$, and $\mathcal{L}_{\rm{LPIPS}}$ yields robust reconstruction and achieves the best overall visual quality. Furthermore, noting that the VQAE and diffusion UNet models are trained separately, the alignment of these two models is also important. As shown in Figure~\ref{fig: ablation ILB regularization loss}, over-optimizing the image latent representation $z_0$ using only the consistency loss can lead to severe deviation from the diffusion latent space and consequently introduce significant distortion. The regularization loss $\mathcal{L}_{\rm{reg}}$ acts as a penalty term that prevents the optimized $z_0$ from deviating excessively from the diffusion latent space. Moreover, as presented in Table~\ref{tab: ablation ILB loss} and Figure~\ref{fig: ablation ILB regularization loss}, introducing $\mathcal{L}_{\rm{reg}}$ improves the robustness of diffusion inversion reconstruction while preserving the reconstruction consistency.

\subsection{Performance in Downstream Tasks}

\begin{table}[t]
\centering
\caption{Comparison results on the image editing task.}
\label{tab: editing quantitative results}
\resizebox{\linewidth}{!}{
\begin{tabular}{lccccc}
\toprule
\multirow{2}{*}{\textbf{\begin{tabular}[c]{@{}c@{}}Inversion\\ Method\end{tabular}}} & \multirow{2}{*}{\textbf{\begin{tabular}[c]{@{}c@{}}Structure\\ Distance$\downarrow$\end{tabular}}} & \multicolumn{2}{c}{\textbf{Unedited Fidelity}} & \multicolumn{2}{c}{\textbf{CLIP Similarity}} \\
 & & \textbf{PSNR} $\uparrow$ & \textbf{SSIM} $\uparrow$ & \textbf{Whole} $\uparrow$ & \textbf{Edited} $\uparrow$ \\
\midrule
DDIM Inv. & 0.0705 & 17.82 & 0.7141 & \textbf{25.18} & \textbf{22.35} \\
FPI & 0.0222 & 23.57 & 0.8109 & 24.33 & 21.57 \\
AIDI & 0.0220 & 23.65 & 0.8112 & 24.33 & 21.56 \\
NTI & \textbf{0.0176} & \textbf{25.87} & \textbf{0.8396} & 24.85 & 22.01 \\
ExactDPM & 0.0336 & 22.55 & 0.7922 & 24.52 & 21.74 \\
LBI-N (ours) & \underline{0.0184} & \underline{25.34} & \underline{0.8332} & \underline{25.06} & \underline{22.24} \\
\bottomrule
\end{tabular}}
\end{table}

\begin{table}[t]
\centering
\caption{Quantitative results on rare concept generation across five rare concepts from the ImageNet-1K dataset.}
\label{tab: rare concept generation quantitative results}
\resizebox{\linewidth}{!}{
\begin{tabular}{lccccc}
\toprule
\textbf{Method} & \textbf{C1} & \textbf{C2} & \textbf{C3} & \textbf{C4} & \textbf{C5} \\
\midrule
SD1 & 0.8521 & 0.3852 & 0.7033 & 0.5083 & 0.6463 \\
NAO-SS & 0.8847 & 0.5143 & 0.7120 & 0.5304 & 0.6610 \\
NAO-SS (w/ LBI) & \textbf{0.8923} & \textbf{0.5366} & \textbf{0.7502} & \textbf{0.5869} & \textbf{0.6977} \\
\bottomrule
\end{tabular}
}
\end{table}

\begin{figure*}
\centering
\includegraphics[width=\linewidth]{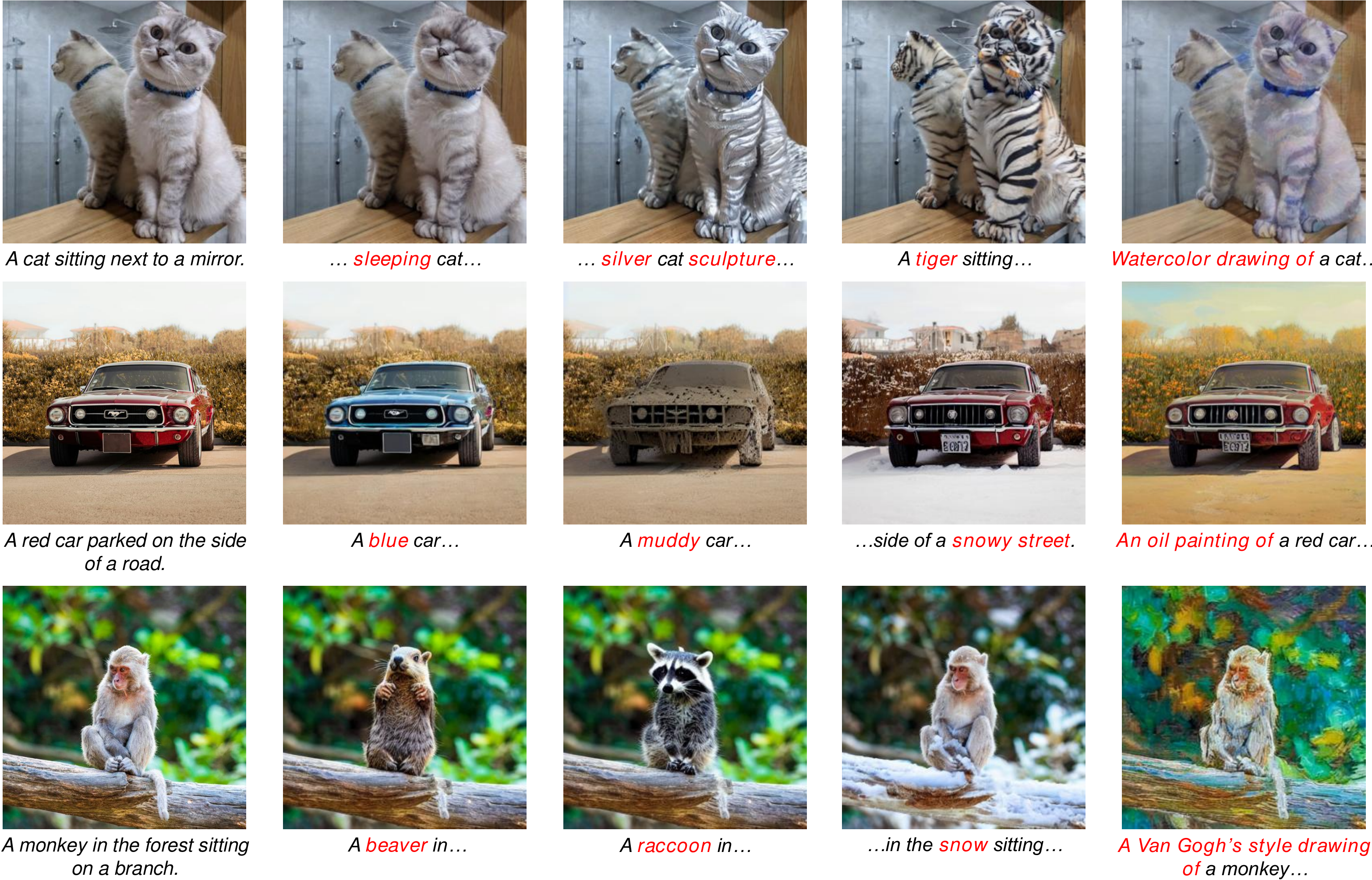}
\caption{Qualitative results of image editing using P2P integrating with the proposed LBI method.}
\label{fig: image editing qualitative}
\end{figure*}

\begin{figure}[t]
\centering
\includegraphics[width=\linewidth]{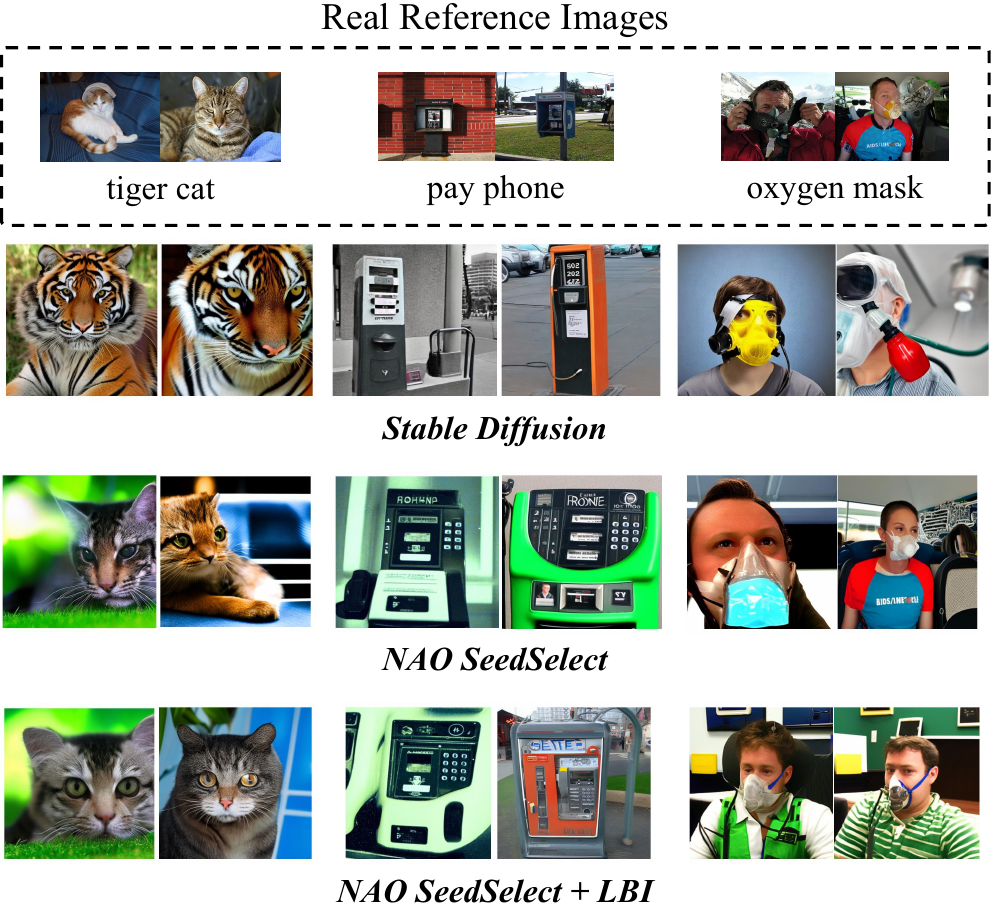}
\caption{Qualitative results of the rare concept generation task. We present the performance for three additional rare concepts: \textit{tiger cat}, \textit{pay phone}, and \textit{oxygen mask}.}
\label{fig: rare concept generation qualitative}
\end{figure}

\subsubsection{Image Editing Task}

We conduct a comprehensive evaluation of image editing on the PIE-Bench dataset, which consists of 700 images across 10 editing categories. We compare our method with DDIM Inversion, FPI, AIDI, NTI, and ExactDPM, and assess the trade-off between fidelity and editability using structural distance, PSNR, SSIM, and CLIP score. Since the inversion results reported in Table~\ref{tab: comparisons image reconstruction} have already demonstrated the robustness of all variants of the proposed method, we adopt only the numerical variant (LBI-N) here due to its superior computational efficiency. The quantitative results are presented in Table~\ref{tab: editing quantitative results}, where the best results are highlighted in bold and the second-best results are underlined. Although the proposed method does not achieve the best performance in terms of either fidelity or editability individually, it attains the best overall trade-off between them. The qualitative results of various editing cases using P2P integrated with our proposed LBI method are shown in Figure~\ref{fig: image editing qualitative}. We consider a diverse set of editing scenarios, including object replacement, color modification, state transformation, material alteration, and style transfer. These results further demonstrate the potential of our method for consistent editing of real-world images.

\subsubsection{Rare Concept Generation Task}

We conduct a quantitative evaluation of rare concept generation by measuring the similarity between reference and generated images using the CLIP image encoder. We compare the performance of SD1 and NAO-SeedSelect (NAO-SS) across five rare concepts from the ImageNet-1K dataset. The quantitative results are presented in Table~\ref{tab: rare concept generation quantitative results}, where the selected rare concepts include B\&W warbler (C1), T-shirt (C2), Space heater (C3), Sweatshirt (C4), and Knife (C5). The results indicate that the proposed LBI method serves as a better seed noise generator, thereby producing generated images that are better aligned with the reference images. In addition to the five concepts reported in Table~\ref{tab: rare concept generation quantitative results}, we present three additional rare concepts in Figure~\ref{fig: rare concept generation qualitative}. The qualitative results indicate that our proposed method enables the model to generate rare concept images with enhanced realism.

\section{Conclusion}

In this work, we study the problem of reconstructing real-world images using diffusion models. We identify two major challenges: (1) the misalignment between the inversion and generation trajectories during the diffusion inversion process, and (2) the mismatch between the diffusion inversion process and the VQAE reconstruction. To address the first challenge, we propose the Latent Bias Optimization (LBO) strategy, which learns a latent bias vector at each inversion step to align the inversion and generation trajectories. We further explore three optimization schemes, \textit{i.e.}, gradient-based, numerical, and hybrid approaches. All variants demonstrate robust inversion performance, with the numerical method being the most computationally efficient. To address the second challenge, we propose the Image Latent Boosting (ILB) technique, which performs joint optimization of the diffusion inversion and VQAE reconstruction processes. Extensive experimental results demonstrate that the proposed method significantly improves the reconstruction quality of diffusion models, while also enhancing performance on downstream tasks, including image editing and rare concept generation.

\section*{Acknowledgments}

This work is supported by the National Natural Science Foundation of China (No. 62331014 and No. 12426312). We also acknowledge the computational support of the Center for Computational Science and Engineering at Southern University of Science and Technology.


\vspace{11pt}

\begin{IEEEbiography}
[{\includegraphics[width=1in,height=1.25in,clip,keepaspectratio]{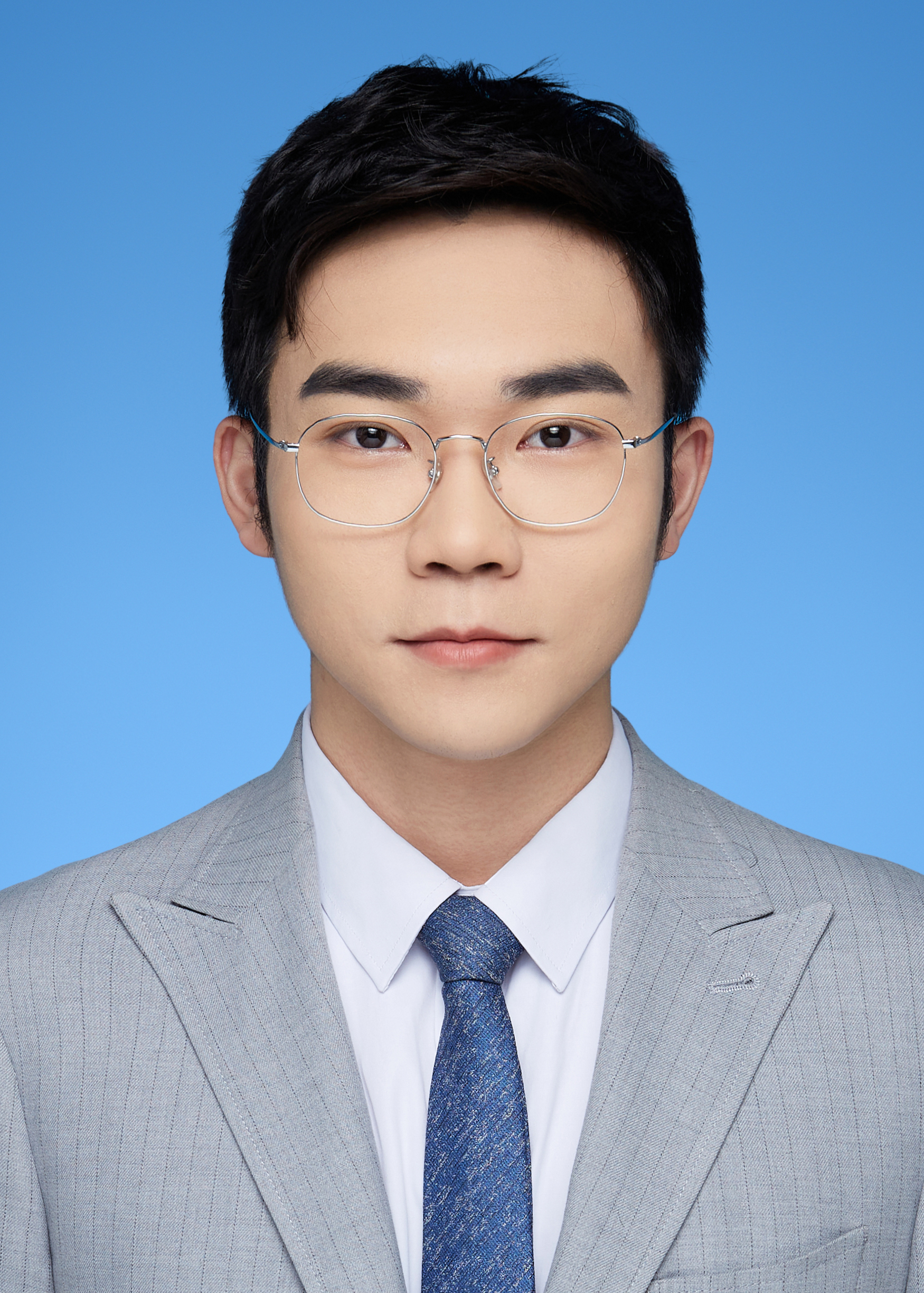}}]{Weiming Chen}
received the B.Eng. degree in mechanical design manufacture and automation, and the M.Sc. degree in electronic science and technology from Xidian University, Xi'an, China, in 2019, and 2023. He is currently pursuing the Ph.D. degree in Intelligent Manufacturing and Robotics from Southern University of Science and Technology, Shenzhen, China. His research interests include machine learning, computer vision, object detection, multimodality, and controllable text-to-image generation.
\end{IEEEbiography}

\begin{IEEEbiography}
[{\includegraphics[width=1in,height=1.25in,clip,keepaspectratio]{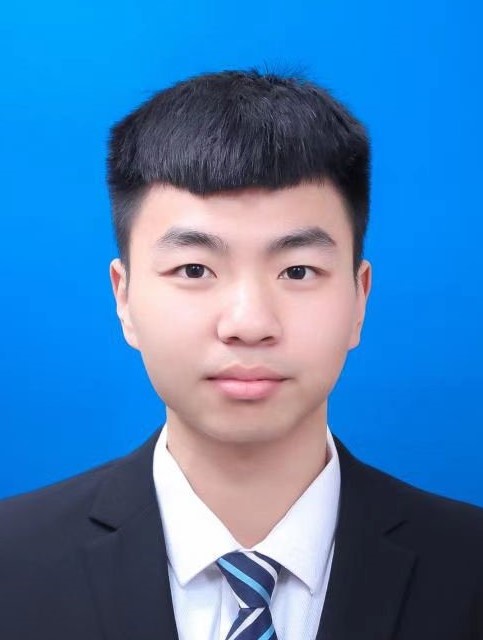}}]{Qifan Liu} received the B.Eng degree in Information and Communication Engineering, Gannan Normal University, Jiangxi, China, in 2017. He received the M.Sc. degree in Integrated Circuit Engineering and Doctor degree in Information and Communication Engineering from Shenzhen University, Shenzhen, China, in 2020 and 2023. He is currently a postdoctoral researcher in the Department of Electrical and Electronic Engineering at Southern University of Science and Technology, Shenzhen, China. His research interests include few-shot learning and visual language models.
\end{IEEEbiography}

\begin{IEEEbiography}
[{\includegraphics[width=1in,height=1.25in,clip,keepaspectratio]{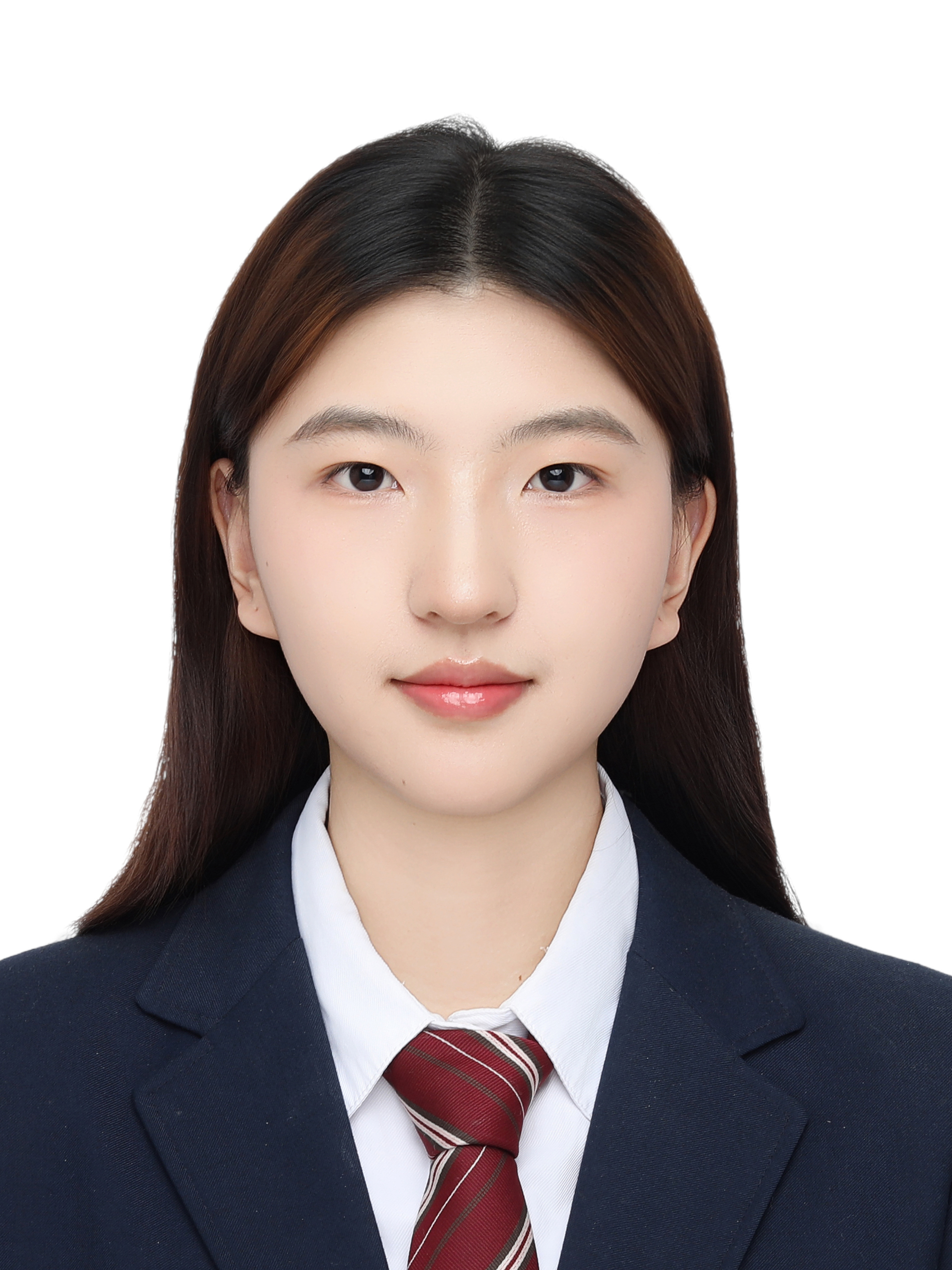}}]{Siyi Liu}
received the B. Eng. degree in Communication Engineering from Shenzhen University, Shenzhen, China, in 2023. She is currently pursuing the M.Sc. degree in Electronic Science and Technology at Southern University of Science and Technology, Shenzhen, China. Her research interests include computer vision, multi-modality, diffusion models, text-to-image generation, and image editing.
\end{IEEEbiography}

\begin{IEEEbiography}[{\includegraphics[width=1in,height=1.25in,clip,keepaspectratio]{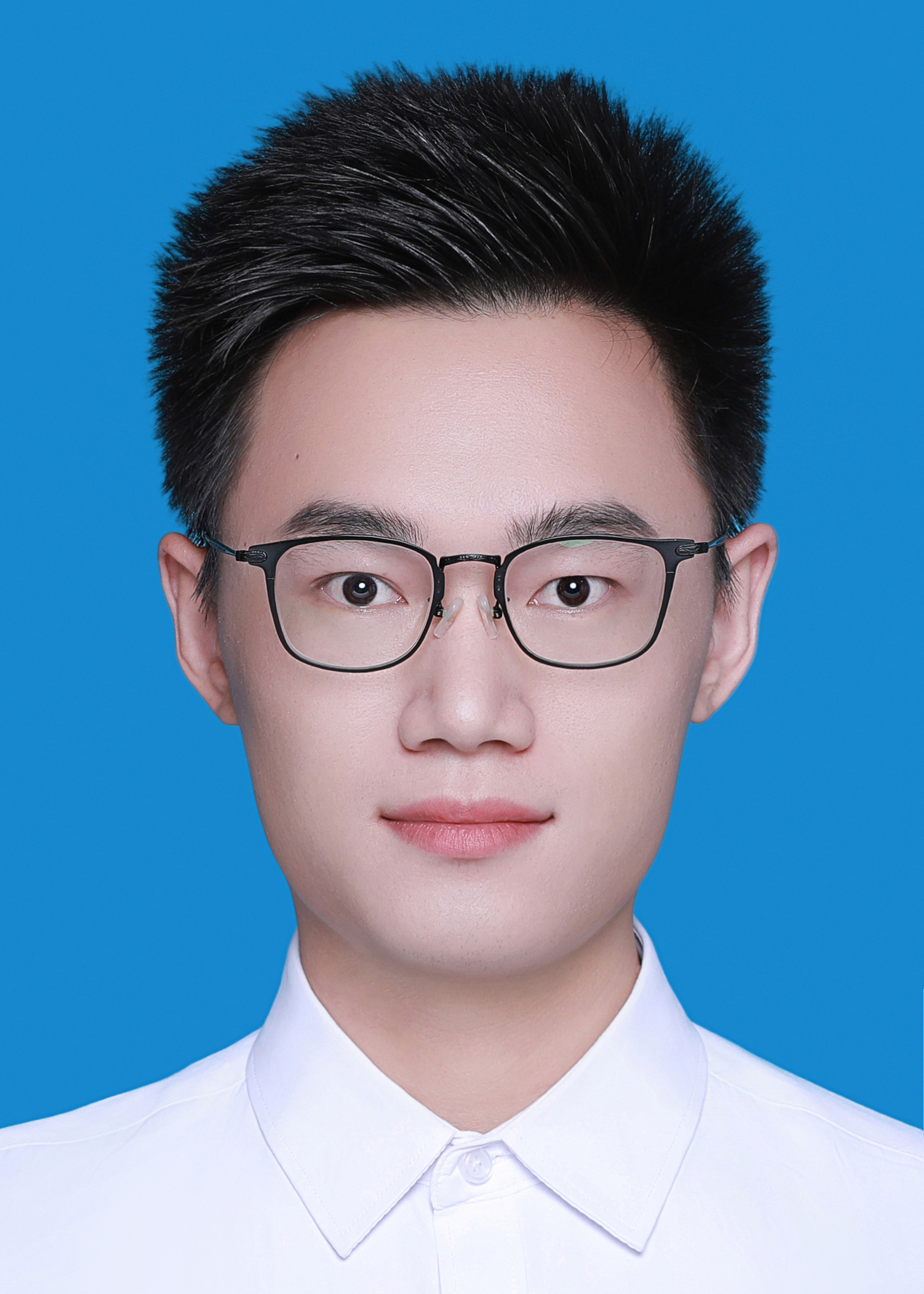}}]{Yushun Tang}
is currently pursuing a Ph.D. in Intelligent Manufacturing and Robotics at the Southern University of Science and Technology (SUSTech). He holds a Bachelor’s degree in Optoelectronic Information Science and Engineering from Harbin Engineering University and a Master’s degree in Electronic Science and Technology from SUSTech. His current research focuses on Computer Vision, Transfer Learning, Domain Adaptation, and Text-to-Image Generation.
\end{IEEEbiography}

\begin{IEEEbiography}
[{\includegraphics[width=1in,height=1.25in,clip,keepaspectratio]{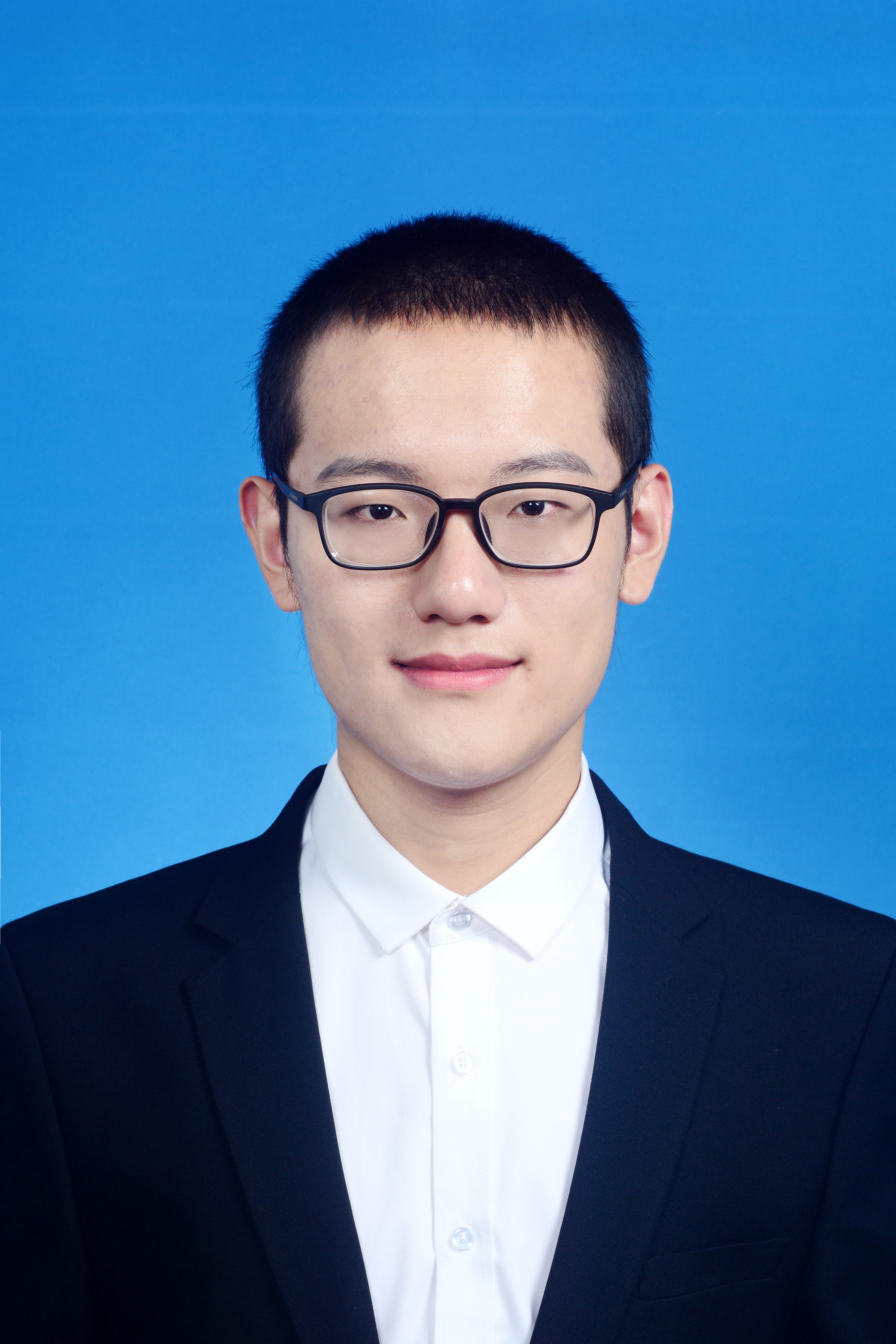}}]{Yijia Wang} is currently pursuing the B.Eng. degree in Electronic and Electrical Engineering from the Southern University of Science and Technology, Shenzhen, China. His research interests include machine learning, image generation models, and computer vision.
\end{IEEEbiography}

\begin{IEEEbiography}
[{\includegraphics[width=1in,height=1.25in,clip,keepaspectratio]{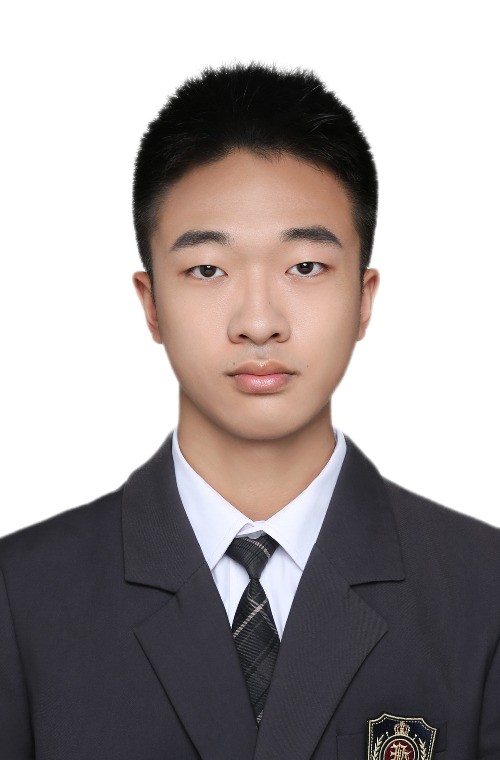}}]{Zhihan Zhu} is currently pursuing the B.Eng. degree in Electronic and Electrical Engineering from the Southern University of Science and Technology, Shenzhen, China. His research interests include machine learning, computer vision, and diffusion models.
\end{IEEEbiography}

\begin{IEEEbiography}
[{\includegraphics[width=1in,height=1.25in,clip,keepaspectratio]{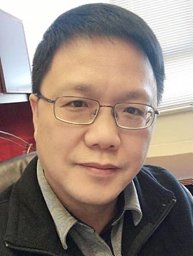}}]{Zhihai He} (Fellow, IEEE) received the B.S. degree in mathematics from Beijing Normal University, Beijing, China, in 1994, the M.S. degree in mathematics from the Institute of Computational Mathematics, Chinese Academy of Sciences, Beijing, China, in 1997, and the Ph.D. degree in electrical engineering from the University of California, at Santa Barbara, CA, USA, in 2001. In 2001, he joined Sarnoff Corporation, Princeton, NJ, USA, as a member of technical staff. In 2003, he joined the Department of Electrical and Computer Engineering, University of Missouri, Columbia, MO, USA, where he was a tenured full professor. He is currently a chair professor with the Department of Electrical and Electronic Engineering, Southern University of Science and Technology, Shenzhen, P. R. China. His current research interests include image/video processing and compression, wireless sensor network, computer vision, and cyber-physical systems.

He is a member of the Visual Signal Processing and Communication Technical Committee of the IEEE Circuits and Systems Society. He serves as a technical program committee member or a session chair of a number of international conferences. He was a recipient of the 2002 {\sc IEEE Transactions on Circuits and Systems for Video Technology} Best Paper Award and the SPIE VCIP Young Investigator Award in 2004. He was the co-chair of the 2007 International Symposium on Multimedia Over Wireless in Hawaii. He has served as an Associate Editor for the {\sc IEEE Transactions on Circuits and Systems for Video Technology} (TCSVT), the {\sc IEEE Transactions on Multimedia} (TMM), and the Journal of Visual Communication and Image Representation. He was also the Guest Editor for the IEEE TCSVT Special Issue on Video Surveillance.
\end{IEEEbiography}

\vfill

\end{document}